\documentclass[conference]{IEEEtran}
\IEEEoverridecommandlockouts
\usepackage{cite}
\usepackage{amsmath,amssymb,amsfonts}
\usepackage{newtxtext}
\usepackage{newtxmath}
\usepackage{algorithmic}
\usepackage{graphicx}
\usepackage{textcomp}
\usepackage{xcolor}
\usepackage{tabularx}
\usepackage{booktabs}
\usepackage{makecell}
\def\BibTeX{{\rm B\kern-.05em{\sc i\kern-.025em b}\kern-.08em
    T\kern-.1667em\lower.7ex\hbox{E}\kern-.125emX}}

\begin{document}

\title{From Phase Transition to Systemic Failure: A Decoupled Analytics Framework for GNN Robustness\\
\thanks{This research was funded by Science and Technology Projects of Xizang Autonomous Region, China, project title "Research and Application of LLM-based Intelligent Tourism Service System for Xizang", grant number XZ202401ZY0008.}
}

\author{
\parbox{\textwidth}{
\centering
\begin{minipage}[t]{0.32\textwidth}
\centering
1\textsuperscript{st} Shuai Yan \\
\textit{College of Computer and Software} \\
\textit{Chengdu Jincheng College}\\
Chengdu 611731, China \\
yanshuai1@cdjcc.edu.cn \\
\end{minipage}
\hspace{0.10\textwidth}
\begin{minipage}[t]{0.32\textwidth}
\centering
2\textsuperscript{nd} Dan Peng \\
\textit{College of Computer and Software} \\
\textit{Chengdu Jincheng College}\\
Chengdu 611731, China \\
pengdan93@cdjcc.edu.cn
\end{minipage}

\vspace{1.2em}

\begin{minipage}[t]{0.32\textwidth}
\centering
3\textsuperscript{rd} Jie Li \\
\textit{College of Computer and Software} \\
\textit{Chengdu Jincheng College}\\
Chengdu 611731, China \\
lijie5787@cdjcc.edu.cn
\end{minipage}
\hfill
\begin{minipage}[t]{0.32\textwidth}
\centering
4\textsuperscript{th} Xiaodong Huang \\
\textit{College of Computer and Software} \\
\textit{Chengdu Jincheng College}\\
Chengdu 611731, China \\
huangxiaodong@cdjcc.edu.cn
\end{minipage}
\hfill
\begin{minipage}[t]{0.32\textwidth}
\centering
5\textsuperscript{th} Ke Wang* \\
\textit{College of Computer and Software} \\
\textit{Chengdu Jincheng College}\\
Chengdu 611731, China \\
wangke@cdjcc.edu.cn
*Corresponding author
\end{minipage}
}
}

\maketitle

\begin{abstract}
Data quality is a major bottleneck for the reliable deployment of graph neural networks (GNNs) in real-world graph mining tasks. Among various sources of degradation, label noise and feature distribution shift (hereafter referred to as distribution shift) are two common yet fundamentally different challenges. To study their effects under controlled conditions, this paper constructs a synthetic homophilic graph regression benchmark in which the two factors can be manipulated separately. A total of 41 configurations and 410 runs are conducted to evaluate the behavior of representative GNN models under varying noise and shift conditions. The results show two distinct patterns. First, under additive label corruption, performance remains relatively stable over a broad range of noise settings and begins to deteriorate sharply only after an observed transition region around the 50\% noise ratio. Second, under extreme feature distribution shift, all tested models suffer substantial degradation, with test MSE increasing by 48$\times$ to 316$\times$ and correlation dropping by 73\% to 89\%. These findings suggest that, in the present controlled setting, GNNs are considerably more tolerant to moderate label perturbation than to severe distribution mismatch. The study provides a controlled empirical baseline for understanding how data quality affects GNN-based graph mining systems and offers practical implications for deployment-oriented monitoring and model maintenance.
\end{abstract}


\section{Introduction}
Graph neural networks (GNNs) have emerged as the foundational architecture for graph mining and machine learning in big data ecosystems \cite{b1}. However, as these models transition from controlled academic benchmarks to dynamic real-world deployments, a critical vulnerability has been exposed: while algorithmic architectures are highly optimized, deployment reliability is fundamentally dictated by the quality, stability, and integrity of the underlying data. In the context of large-scale data analytics, shifting the focus from model-centric structural upgrades to data-centric robustness evaluation has become an urgent necessity \cite{b1}\cite{b2}.

In dynamic deployment environments, data degradation typically manifests through two pervasive yet mechanistically distinct channels: label noise (inaccurate, incomplete, or corrupted annotations) and feature distribution shift (systematic statistical divergence between the training environment and the deployment phase). Although both phenomena are widely acknowledged in data science, their impact on GNNs remains highly entangled in existing literature. Current out-of-distribution benchmarks often mix heterogeneous real-world datasets where label noise, structural perturbations, and distribution shifts co-occur simultaneously \cite{b2}\cite{b7}\cite{b11}. Consequently, a critical research gap persists: without decoupling these confounding variables, it is impossible to pinpoint the precise failure mechanisms of GNNs. When a graph mining model collapses in production, is it because the supervision signal was corrupted, or because the input data distribution drifted?

This paper addresses this fundamental diagnostic challenge by establishing a controlled empirical data analytics framework. Rather than proposing a more complex GNN variant, we aim to isolate data quality factors and quantitatively examine their independent effects on prediction behavior. To bridge the gap between real-world complexity and experimental control, we construct a fully synthetic homophilic graph regression platform. Crucially, to ensure this synthetic ``sandbox'' reflects realistic data dynamics rather than arbitrary assumptions, we ground our generation process in empirical reality. Specifically, we extract robust statistical priors and calibrate the data generation parameters using a curated collection of 11 real-world ecological observation datasets. This methodology provides a rigorously parameterized virtual testbed where additive label corruption and extreme distribution shifts can be manipulated as strictly independent variables under a unified protocol.

By executing this systematic isolation, we uncover profound implications for big data machine learning deployments. The contributions of this paper are as follows:

(1) A decoupled quantitative data analytics testbed is developed. By leveraging a comprehensive suite of real-world ecological statistics as calibration priors, we construct a fully synthetic graph platform to strictly isolate and evaluate the independent impacts of label noise and distribution shift.

(2) Under the present protocol, the study diagnoses two distinct failure modes in GNNs: a progressive degradation under label noise, with a visible transition interval in the 40\%--60\% range, contrasted by severe degradation under extreme distribution shift.

(3) This vulnerability pattern is demonstrated to be architecture-consistent across representative GNN backbones (GCN and GAT) and capacity scales, suggesting that algorithmic complexity does not substitute for data integrity in the present setting.

(4) The study provides deployment-oriented heuristics for industrial graph mining, revealing that proactive input drift monitoring and recalibration are far more critical safeguards than supervision noise mitigation alone.

The remainder of the paper is organized as follows. Section II reviews prior work on GNN robustness and out-of-distribution generalization on graphs. Section III presents the methodological framework, including the empirical calibration basis, the synthetic sandbox construction, the degradation protocols, and the experimental settings. Section IV reports the quantitative results under both label noise and distribution shift. Section V discusses the practical implications, limitations, and deployment-oriented insights. Finally, Section VI concludes the paper.

\section{Related Work}
\subsection{GNN Robustness under Data Degradation}
Prior studies on GNN robustness have mainly focused on adversarial perturbations, structural attacks, and architecture-related degradation, while progressively increasing non-adversarial label corruption remains less systematically characterized in controlled graph regression settings \cite{b3}\cite{b4}\cite{b5}\cite{b6}. In parallel, graph OOD benchmarks such as WILDS and GOOD have shown that graph models often degrade substantially under domain shift, temporal variation, and environmental heterogeneity. However, the realism of these benchmarks also means that multiple degradation sources are typically entangled. This work complements such benchmarks by constructing a calibrated controlled sandbox in which label noise and feature distribution shift can be manipulated independently, enabling more interpretable failure attribution under a unified protocol.

\subsection{Out-of-Distribution Generalization on Graphs}
A parallel line of research has investigated out-of-distribution (OOD) generalization in graph learning. Large-scale benchmarks such as WILDS and GOOD have played an important role in revealing that graph models often experience substantial performance drops once deployment conditions deviate from the training environment \cite{b7}. Related research on size-invariant graph representations and graph OOD generalization has further expanded empirical understanding of how graph models behave under distribution mismatch \cite{b8}\cite{b9}, while architectural analyses of graph attention mechanisms have improved understanding of attention behavior in graph settings \cite{b10}. These benchmark efforts have significantly advanced empirical understanding of graph OOD evaluation by introducing realistic domain shifts, temporal variation, and cross-environment heterogeneity \cite{b11}.

However, the realism of such benchmarks also introduces a methodological limitation for diagnostic analysis: the underlying sources of degradation are usually entangled. In real-world benchmark graphs, feature shift, structural perturbation, label bias, missingness, and environment-specific sampling artifacts often co-occur. As a result, although these benchmarks are highly valuable for measuring practical OOD difficulty, they are less suitable for isolating the independent causal contribution of a single degradation factor.

This paper complements, rather than replaces, existing OOD benchmarks. Our goal is not to build a more realistic benchmark than GOOD or WILDS, but to construct a calibrated controlled sandbox in which label noise and feature distribution shift can be manipulated as independent variables under a unified protocol. By sacrificing some environmental complexity in exchange for variable isolation, the study provides a more interpretable basis for identifying which data-quality failure mode is primarily responsible for model collapse \cite{b12}\cite{b13}.

\section{Methods and Experimental Design}

\subsection{Empirical Basis and Variable Decoupling Strategy}
The empirical basis of this study consists of 11 real-world ecological observation datasets with 9,643 records collected from Sejila Mountain in Tibet, the Three-River Source region of Qinghai, and grassland sites in Gannan and Haibei during 2007–2023. These datasets include continuous ecological variables such as meteorological conditions, soil properties, vegetation biomass, nutrient content, and leaf area index, together with continuous regression targets. Their statistical properties are used to calibrate realistic feature ranges, target scales, and distributional variability.

Direct evaluation on the raw ecological datasets would confound multiple degradation sources, including measurement error, label corruption, seasonal drift, and spatial heterogeneity. Therefore, rather than training directly on the entangled real-world data, we use them only as an empirical calibration source and construct a controlled synthetic graph sandbox in which label noise and feature shift can be manipulated independently.

The empirical calibration source spans multiple ecologically heterogeneous regions and long-term observation periods, providing a realistic basis for estimating feature ranges, target scales, and cross-variable covariance patterns before constructing the controlled synthetic sandbox.

\subsection{Construction of the Calibrated Synthetic Sandbox}
To execute the precise attribution described above, a calibrated synthetic graph testbed---referred to as the experimental ``sandbox''---is constructed. In this controlled environment, the training graph is standardized to 500 nodes, and the Out-of-Distribution (OOD) test graph contains 200 nodes. While real-world big data graphs are significantly larger, this deliberate scale is a standard practice in controlled robustness analytics; it eliminates the computational and topological noise of massive graphs, allowing us to mathematically isolate the fundamental mechanisms of data degradation.

Each node is represented by a 6-dimensional feature vector drawn from a standardized Gaussian baseline, calibrated by the dimensionality of our empirical priors:
\begin{equation}
x_i\sim\mathcal{N}(0,I)
\end{equation}
Continuous regression labels are generated through a linear mapping mechanism to establish a ground-truth oracle:
\begin{equation}
y_i=x_i^Tw+\epsilon_i
\end{equation}
where the weight vector is sampled from $w\sim U(1.0,2.0)$, and the intrinsic observation noise follows $\epsilon_i\sim\mathcal{N}(0,0.05)$. After feature generation, graph topologies are constructed using K-nearest-neighbor (KNN) relationships in the feature space ($K=5$) under Euclidean distance. The resulting homophilic graphs serve as the baseline environment for the node-level regression task, formalized as learning a mapping $f:(X,E)\rightarrow y$.

\subsection{Label Noise Injection Scheme}
To simulate supervisory corruption---such as measurement disturbances or annotation errors in practical data engineering---we introduce additive label noise during the training phase. A proportion $p$ of the node labels is randomly selected and perturbed:
\begin{equation}
y_i^{noise}=y_i+\eta_i
\end{equation}
where $\eta_i$ follows a zero-mean Gaussian distribution. The degradation is systematically controlled via two orthogonal dimensions: the noise ratio (proportion of contaminated samples) and the noise intensity (the variance magnitude of the corruption).

\subsection{Distribution Shift Construction Scheme}
Unlike label noise, which degrades the supervision signal, distribution shift compromises the fundamental identically distributed (i.i.d.) assumption of machine learning. To evaluate OOD generalization, we simulate deployment-time distribution drift caused by spatial transitions or temporal evolution. The training graph remains drawn from the baseline distribution, while the OOD test graph is generated by systematically applying mean vector offsets and covariance matrix scaling:
\begin{equation}
\mu'=\mu+\Delta,\quad \Sigma'=\beta\Sigma
\end{equation}
Under this mathematical formulation, the test features differ systematically from the training set, explicitly testing the model's robustness to input data drift while the underlying regression mechanism remains static.

To instantiate a clearly severe OOD regime, the extreme-shift condition is implemented using a mean offset on the order of 5 standard deviations together with covariance scaling. This choice should not be interpreted as a literal simulation of one specific ecological event. Instead, it serves as a stress-testing configuration that deliberately moves the test distribution far beyond ordinary local fluctuation ranges. The purpose is to construct a deployment-mismatch regime that is unambiguously outside the training support, so that collapse-like failure under severe shift can be distinguished from the more gradual degradation induced by label noise. Because the shift magnitude is calibrated relative to empirical ecological statistics, the resulting sandbox remains anchored in realistic data scale rather than arbitrary synthetic perturbation.

\subsection{Representative GNN Backbones for Controlled Analytics}
This study evaluates two foundational graph neural network architectures: Graph Convolutional Networks (GCN) and Graph Attention Networks (GAT), configured in both standard and large capacity variants \cite{b1}\cite{b10}. The objective of this selection is not to chase state-of-the-art benchmark scores, but to verify whether the identified data vulnerability patterns are architecture-agnostic and consistent across different model complexities \cite{b8}\cite{b9}.

\subsection{Implementation Details}

\begin{table}[htbp]
\centering
\caption{GNN Model Configuration}
\renewcommand{\arraystretch}{1.5}
\resizebox{\columnwidth}{!}{
\begin{tabular}{|l|l|c|c|c|c|c|c|}
\hline
Model & Architecture & Layers & Hidden Dim & Heads & Dropout & BatchNorm & Params \\
\hline
GCN-S & GCN & 2 & 32 & -- & 0 & No & 1K \\
\hline
GCN-L & GCN & 4 & 128 & -- & 0.3 & Yes & 66K \\
\hline
GAT-S & GAT & 2 & 32 & 2 & 0 & No & 5K \\
\hline
GAT-L & GAT & 4 & 128 & 4 & 0.3 & Yes & 530K \\
\hline
\end{tabular}
}
\end{table}

For reproducibility, GCN-S and GAT-S use 2 layers, whereas GCN-L and GAT-L use 4 layers with larger hidden dimensions and stronger regularization. All models are trained with Adam (learning rate 0.01, weight decay 5e-4) for up to 200 epochs with early stopping (patience 30), using mean squared error loss and 10 random seeds.

\subsection{Evaluation Metrics}
To quantify predictive reliability, Mean Squared Error (MSE) and the Pearson correlation coefficient $r$ are jointly utilized. MSE captures the absolute prediction deviation:
\begin{equation}
MSE=\frac{1}{n}\sum_{i=1}^{n}(y_i-\hat{y}_i)^2
\end{equation}
The Pearson correlation coefficient assesses the linear consistency and trend preservation between predictions and ground-truth values:
\begin{equation}
r=\frac{\sum_i(y_i-\bar{y})(\hat{y}_i-\bar{\hat{y}})}{\sqrt{\sum_i(y_i-\bar{y})^2}\sqrt{\sum_i(\hat{y}_i-\bar{\hat{y}})^2}}
\end{equation}
Additionally, Cohen's $d$ is employed to measure effect sizes, and the Wilcoxon rank-sum test is applied to guarantee the statistical significance of the performance degradation \cite{b12}.

\subsection{Experimental Procedure and Statistical Testing}
The analytical framework is divided into two primary diagnostic branches: the label-noise branch and the distribution-shift branch. In the label-noise branch, GCN-S serves as the diagnostic backbone to trace the progressive degradation trajectory cleanly. In the distribution-shift branch, all four model variants (GCN-S, GCN-L, GAT-S, GAT-L) are subjected to both normal (i.i.d.) and extreme (OOD) conditions to test architecture-wide vulnerability.

To ensure statistical rigor and mitigate initialization variance, each configuration is executed 10 times with distinct random seeds, yielding a total of 41 configurations and 410 runs. This comprehensive protocol guarantees that the comparative analytics between label noise and distribution shift are highly reproducible and statistically robust.

\section{Results}

\subsection{Model Behavior Analytics under Label Noise: The Phase Transition}
Under corrupted supervision, the model exhibits a predictable, progressive degradation trajectory rather than an immediate collapse. Following the experimental design, the label-noise experiments fix GCN-S as the representative model and jointly vary the noise ratio and noise intensity to evaluate model sensitivity to degraded supervision. In the current setting, the noise ratio increases from 0\% to 100\%, while the noise intensity is set to $\alpha\in\{0.5,1.0,1.5\}$. 

Our analytics reveal that the model exhibits a certain degree of stability under low-to-moderate label noise. Taking the representative results under a noise intensity of $1.5\times\mathrm{std}$ as an example, when the noise ratio increases from 0\% to 30\%, the test MSE rises only slightly from $3.49\pm1.01$ to $3.65\pm0.97$, while the Pearson correlation coefficient decreases marginally from $0.873\pm0.019$ to $0.862\pm0.020$. In this stage, the performance change remains relatively mild, indicating that under a limited level of supervisory disturbance, neighborhood aggregation over the graph structure is still able to buffer local errors and maintain reasonably stable predictive performance \cite{b5}\cite{b6}.

\begin{figure}[!t]
\centering
\includegraphics[width=0.8\linewidth]{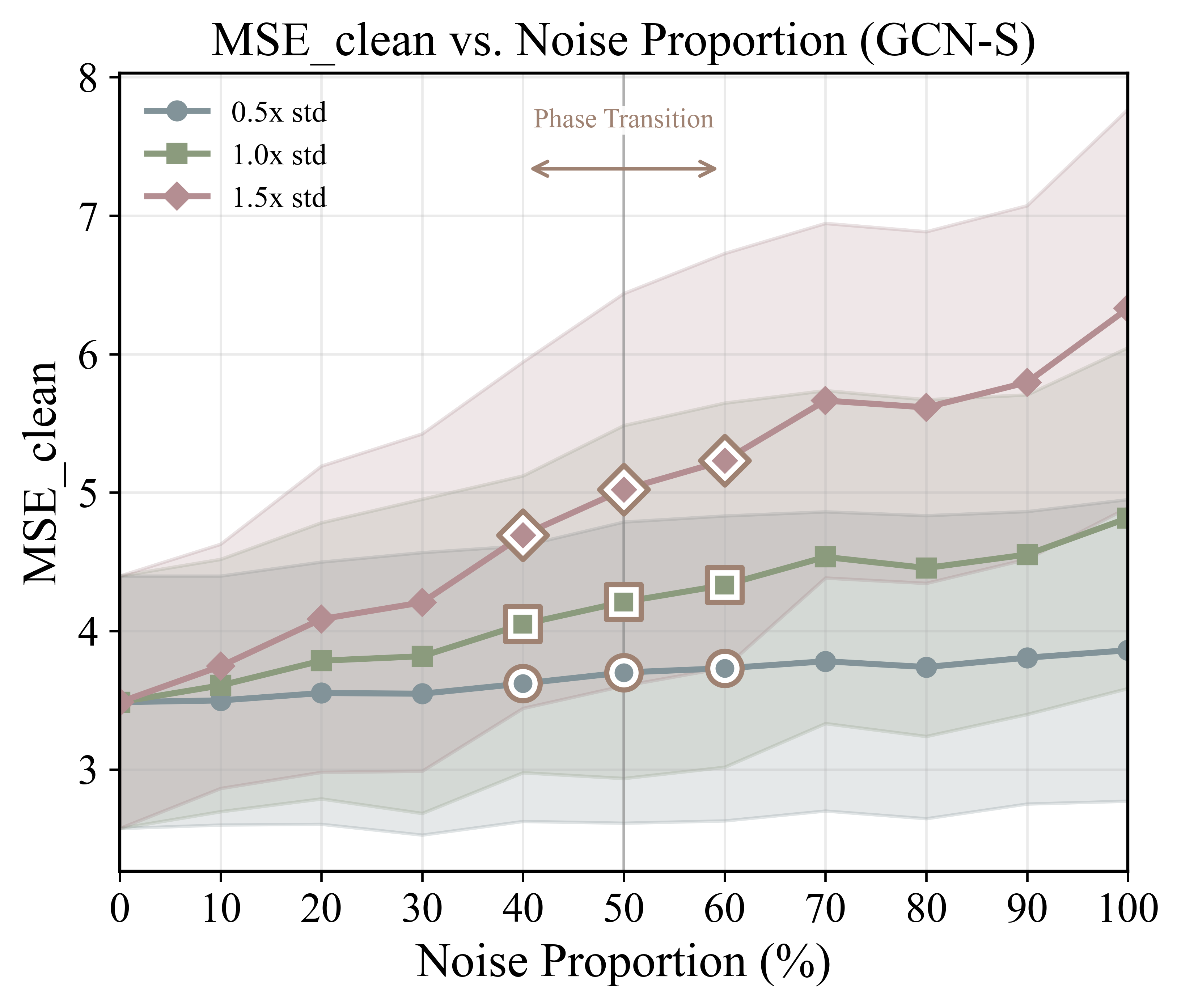}
\caption{MSE trend under increasing noise ratio}
\label{fig:noise_mse}
\end{figure}

\begin{figure}[!t]
\centering
\includegraphics[width=0.8\linewidth]{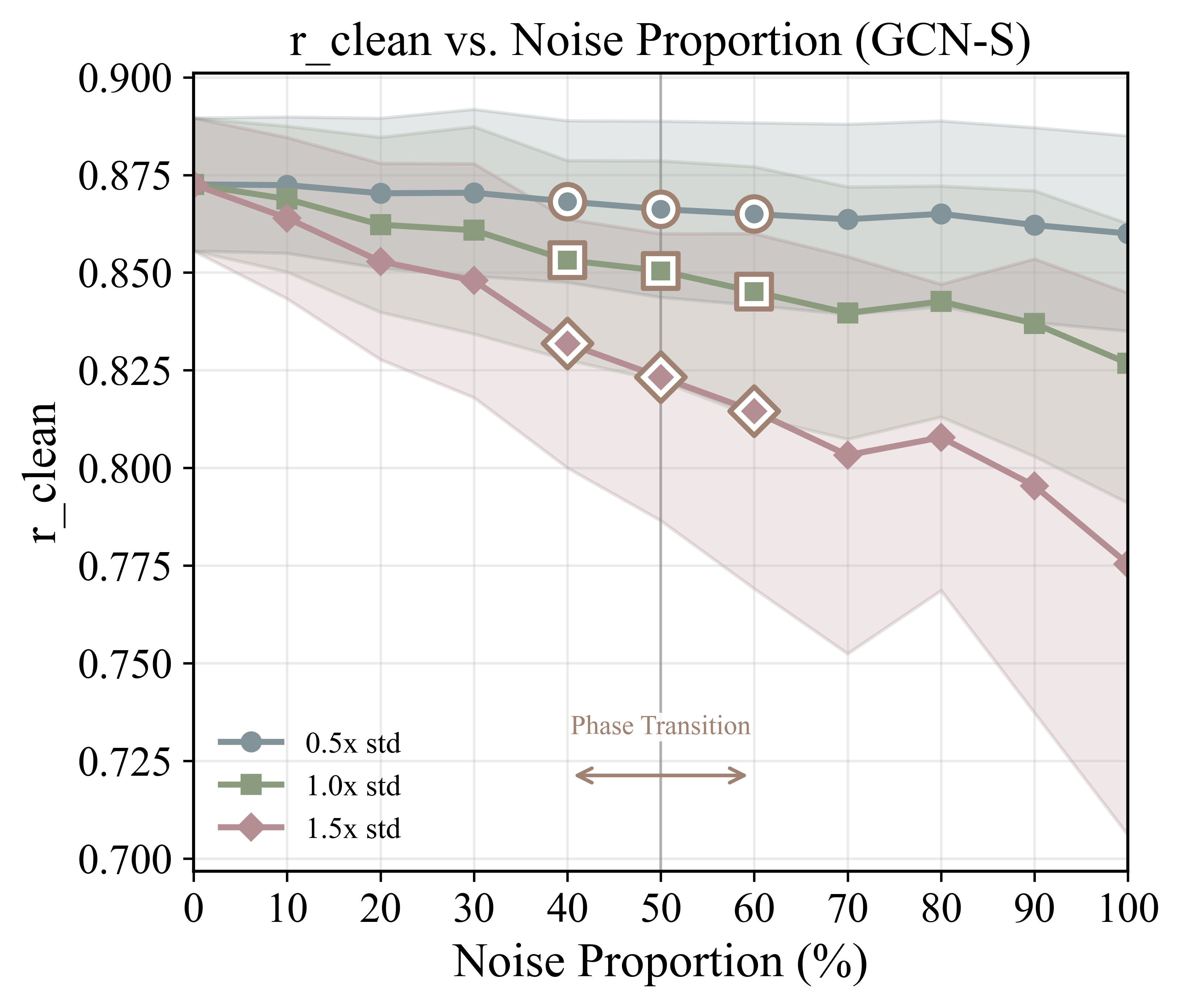}
\caption{Correlation trend under increasing noise ratio}
\label{fig:noise_corr}
\end{figure}

Higher noise ratios and intensities consistently accelerate performance degradation (Fig.~\ref{fig:noise_mse}, \ref{fig:noise_corr}). Notably, this deterioration is gradual rather than abrupt. Under the $1.5\times\mathrm{std}$ setting, the 40\%--60\% interval exhibits a smooth transition: test MSE rises progressively ($4.69\pm1.25 \rightarrow 5.02\pm1.42 \rightarrow 5.23\pm1.50$), while Pearson $r$ steadily declines ($0.832\pm0.032 \rightarrow 0.823\pm0.037 \rightarrow 0.815\pm0.045$). This confirms a gradual transition around 50\% instead of a single-point collapse.

These results confirm that while GNNs are highly fault-tolerant to moderate annotation noise, widespread supervision corruption will eventually exceed the robustness boundary of the model. It should be noted that the observed phase transition around 50\% should be interpreted as an empirical phenomenon under the present analytics setting, rather than a fixed universal threshold.

\subsection{OOD Generalization under Distribution Shift: The Capacity Illusion}
While label noise degrades the target mapping, distribution shift systematically violates the fundamental identically distributed (i.i.d.) assumption of the input space. To evaluate Out-of-Distribution (OOD) vulnerability, four model variants (GCN-S, GCN-L, GAT-S, and GAT-L) are subjected to extreme feature drifts, mathematically defined by a $5\sigma$ mean offset coupled with covariance scaling.

\begin{table}[htbp]
\centering
\caption{Cross-Model Performance under Normal vs. Extreme Distribution Shift}
\label{tab:shift_results}
\renewcommand{\arraystretch}{1.25}
\resizebox{\columnwidth}{!}{
\begin{tabular}{|l|l|l|l|l|l|l|}
\hline
Model & Condition & Test MSE & Test $r$ & MSE Ratio & $r$ Degradation & Cohen's $d$ \\
\hline
GCN-S & Normal  & 2.77$\pm$1.49 & 0.808$\pm$0.022 & -- & -- & -- \\
\hline
GCN-S & Extreme & 691$\pm$759 & 0.178$\pm$0.217 & 249$\times$ & 78\% & 5.15 \\
\hline
GCN-L & Normal  & 3.45$\pm$1.70 & 0.751$\pm$0.051 & -- & -- & -- \\
\hline
GCN-L & Extreme & 457$\pm$452 & 0.085$\pm$0.194 & 133$\times$ & 89\% & 4.74 \\
\hline
GAT-S & Normal  & 2.54$\pm$1.68 & 0.838$\pm$0.050 & -- & -- & -- \\
\hline
GAT-S & Extreme & 805$\pm$904 & 0.229$\pm$0.241 & 316$\times$ & 73\% & 4.51 \\
\hline
GAT-L & Normal  & 3.42$\pm$2.07 & 0.803$\pm$0.039 & -- & -- & -- \\
\hline
GAT-L & Extreme & 165$\pm$123 & 0.212$\pm$0.218 & 48$\times$ & 74\% & 3.68 \\
\hline
\end{tabular}
}
\end{table}

\begin{figure}[!t]
\centering
\includegraphics[width=0.9\linewidth]{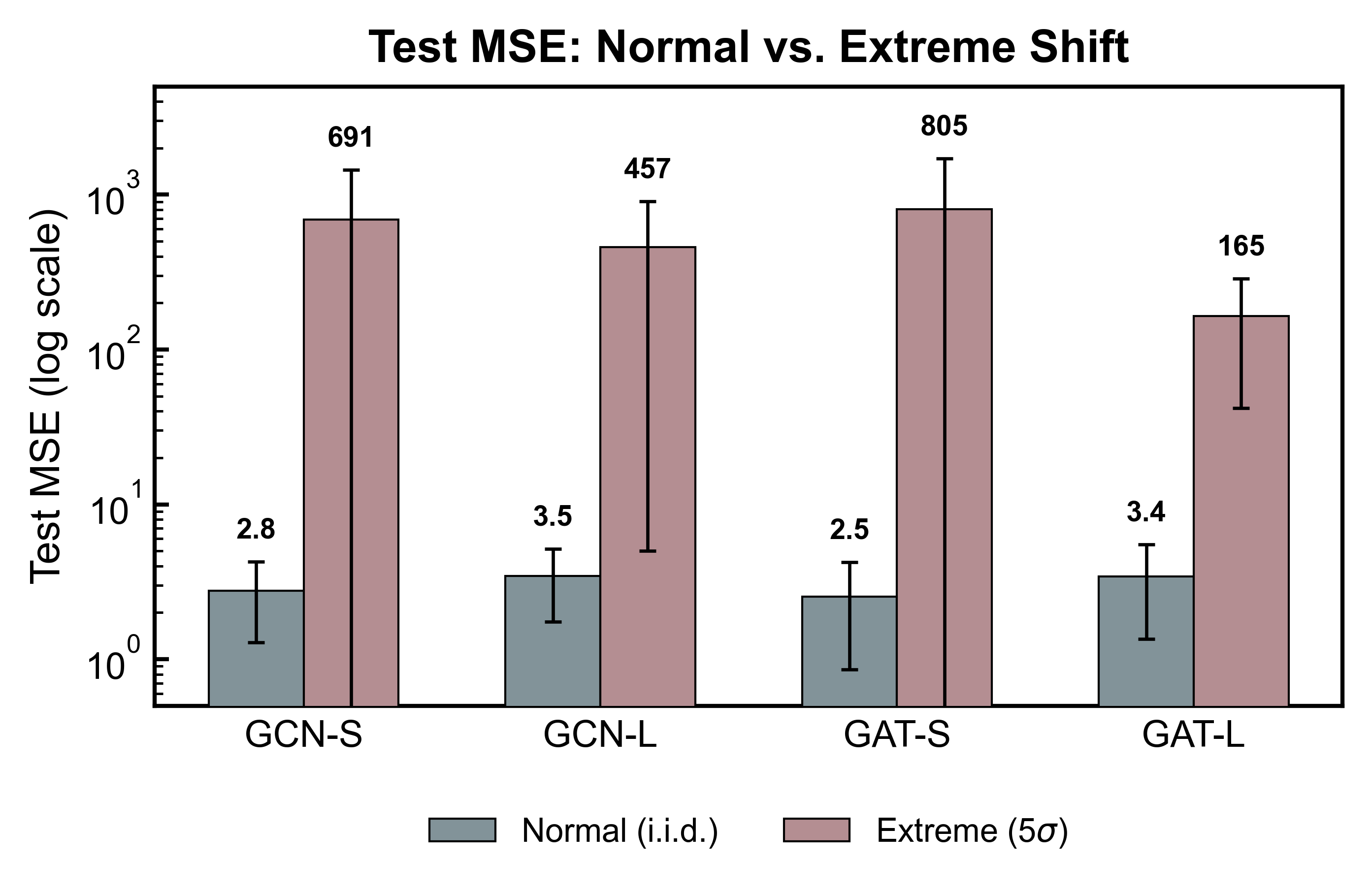}
\caption{MSE change under normal vs. shifted distributions}
\label{fig:shift_mse}
\end{figure}

\begin{figure}[!t]
\centering
\includegraphics[width=0.9\linewidth]{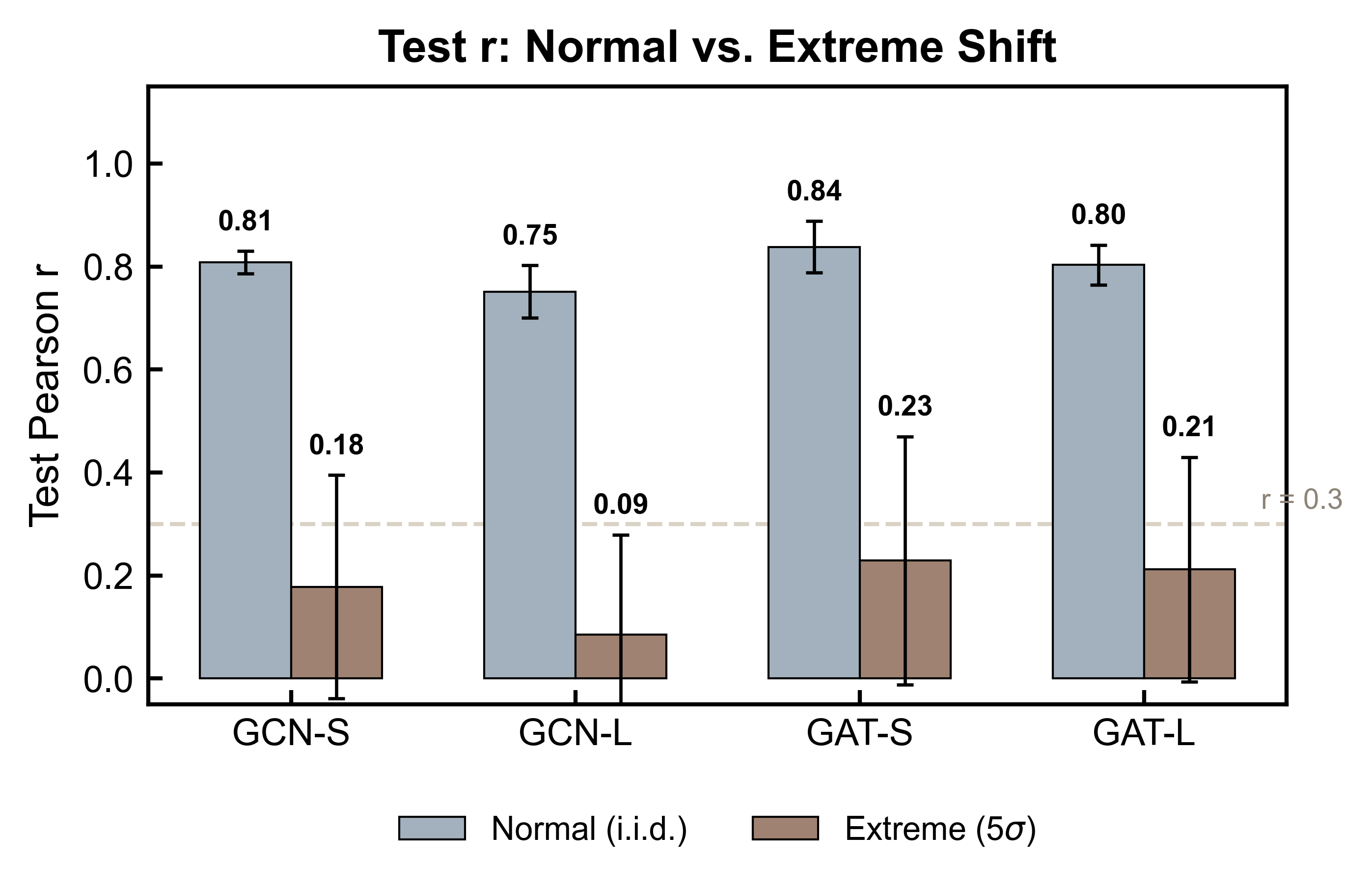}
\caption{Correlation degradation under normal vs. shifted distributions}
\label{fig:shift_corr}
\end{figure}

\begin{figure}[!t]
\centering
\includegraphics[width=0.9\linewidth]{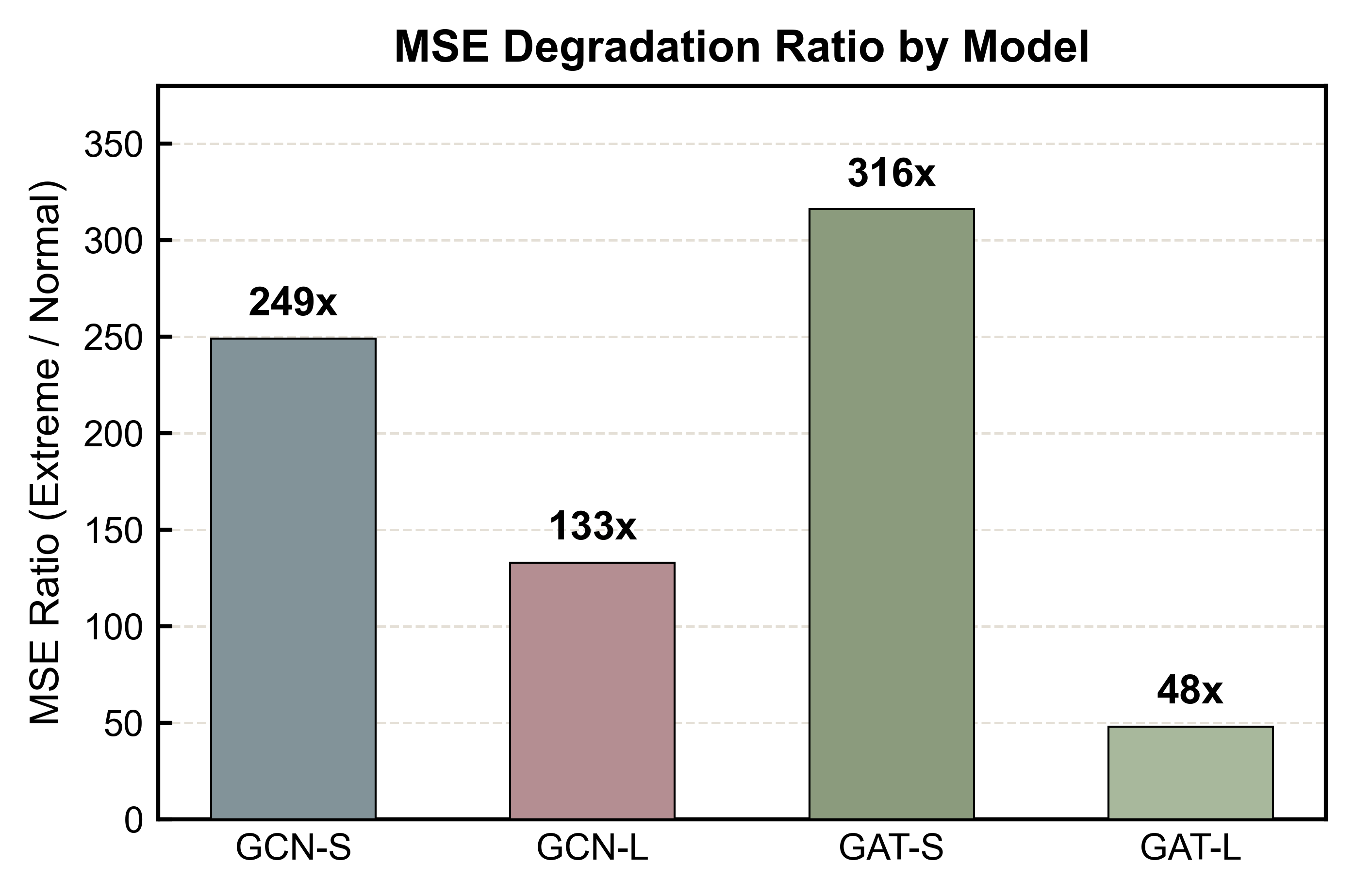}
\caption{Prediction stability under distribution shift}
\label{fig:shift_stability}
\end{figure}

Our analytics demonstrate that distribution shift inflicts catastrophic, collapse-like failures that drastically eclipse the damage caused by label noise. Under the normal condition, models maintain relatively good predictive performance. However, under extreme shift, all models suffer severe performance collapse. For a lightweight model (GCN-S), MSE increases from $2.77\pm1.49$ to $691\pm759$ (about $249\times$), while correlation drops from $0.808\pm0.022$ to $0.178\pm0.217$. Overall, MSE amplification ranges from 48$\times$ to 316$\times$, and Pearson correlation degradation ranges from 73\% to 89\% across the board. The detailed cross-model statistics are summarized in Table~\ref{tab:shift_results}.

More crucially, our cross-model comparison shatters the ``capacity illusion'' in deep graph learning. The high-complexity GAT-L variant demonstrates exceptional in-distribution (i.i.d.) fitting capabilities; nevertheless, under OOD conditions, its test MSE still surges to $165\pm123$ ($48\times$), accompanied by a severe 74\% correlation collapse. This pronounced separation, highlighted in Fig.~\ref{fig:shift_stability}, indicates that stronger in-distribution expressiveness does not necessarily translate to better OOD robustness. Increased parameter complexity does not compensate for severe underlying distribution mismatch.

To ensure these findings are strictly empirical rather than artifacts of stochastic initialization, statistical rigor is prioritized. The computed Cohen's $d$ values reported in Table~\ref{tab:shift_results} verify exceptionally large effect sizes. This suggests that the observed degradation is systematic rather than attributable to random variation.

\subsection{Comparative Synthesis of Failure Modes}
Taken together, the results of the two experimental branches show that label noise and distribution shift, although both belonging to data quality degradation, influence GNNs in fundamentally different ways. This contrast is consistently supported by the trend curves in Fig.~\ref{fig:noise_mse}--\ref{fig:noise_corr}, the cross-condition comparisons in Fig.~\ref{fig:shift_mse}--\ref{fig:shift_stability}, and the numerical summary in Table~\ref{tab:shift_results}.

Label noise acts as a \textit{progressive attenuator}: it weakens the reliability of the supervisory signal, but can be partially buffered by the graph's structural smoothing until it reaches the transition threshold. Conversely, distribution shift acts as a \textit{systematic destroyer}: it abruptly undermines the validity of the input feature space upon which the learned mapping inherently relies.

\section{Discussion}
\subsection{The Low-Pass Filtering Effect of Graph Topology}
Under the current controlled analytics framework, label noise produces a progressive degradation trajectory. A theoretical explanation, rooted in graph signal processing, is that neighborhood aggregation in homophilic graphs inherently acts as a structural low-pass filter \cite{b5}. When annotation corruption remains low-to-moderate, the message-passing mechanism effectively smooths out independent label noise by aggregating signals from partially informative surrounding nodes. This topological buffering explains the model's initial resilience. However, this capacity is strictly bounded; once the contamination ratio surpasses the critical phase transition threshold, the aggregated representations become predominantly corrupted, leading to the rapid performance decay observed in our analytics \cite{b4}\cite{b6}.

\subsection{The Global Vulnerability to Feature Drift}
In contrast to local label corruption, distribution shift presents a systemic failure mode. It does not merely weaken the supervision signal; it fundamentally alters the input statistics and violates the core identically distributed (i.i.d.) assumption of the deployment environment. In this context, distribution shift drastically displaces the valid operating manifold of the learned mapping. When models are forced to extrapolate on out-of-distribution (OOD) test data, the neighborhood aggregation mechanism provides no defensive utility against global feature drift, resulting in the abrupt, collapse-like prediction errors across all tested architectures \cite{b7}\cite{b9}\cite{b13}.

\subsection{Deployment Implications for MLOps}
Our comparative analytics explicitly shatter the assumption that scaling up model complexity can unilaterally circumvent data quality vulnerabilities. For industrial-scale graph mining and Machine Learning Operations (MLOps) pipelines, these findings dictate a clear risk management hierarchy and a paradigm shift in resource allocation.

While moderate label noise can often be mitigated downstream via data cleaning or lightweight training strategies, distribution shift represents an existential deployment risk. Rather than solely investing in post-hoc label correction or engineering heavier model architectures, system deployments must prioritize \textit{Input Distribution Drift Monitoring}. Establishing automated detection thresholds and triggering continuous recalibration workflows the moment feature drift is detected are indispensable strategies for maintaining system reliability and preventing catastrophic baseline failures \cite{b11}.

\subsection{Limitations}
This study employs a calibrated synthetic graph platform to prioritize exact variable isolation over unconstrained real-world complexity. By anchoring the data generation parameters solely in real ecological statistics, we ensured the empirical validity of our testbed. Nonetheless, these conclusions serve as controlled baseline analytics, and certain limitations remain. Specifically, the distribution-shift branch currently reports only the baseline and a severe extreme-shift condition, but does not separately present intermediate shift gradients, which slightly constrains finer-grained practical risk assessments. Future research should not only address these intermediate shift levels but also transition from this controlled sandbox to massive, non-stationary real-world networks to further validate these robustness boundaries under deeply entangled degradation factors \cite{b7}\cite{b11}.

\section{Conclusion}
This paper establishes a rigorous, controlled empirical data analytics framework to isolate and compare the impacts of two pervasive data quality challenges in graph mining: label noise and distribution shift. By deploying a synthetic testbed strictly calibrated with real-world ecological statistics, we successfully decoupled these confounding variables and identified two fundamentally distinct failure modes. The analytics reveal that GNNs exhibit topological resilience to moderate label noise---acting as a structural low-pass filter until a critical phase transition is reached---but suffer catastrophic, collapse-like failures under extreme distribution shift, regardless of the underlying model capacity.

Ultimately, this study answers a critical deployment question for big data machine learning: to safeguard production systems, algorithmic structural upgrades cannot substitute for data integrity. Industrial graph mining pipelines must strategically pivot toward proactive input feature distribution drift monitoring and lifecycle data maintenance, treating systemic feature shift as a substantially higher-priority operational risk than localized supervision noise.

\end{document}